\documentclass[sigconf]{acmart}

\usepackage{microtype}
\usepackage{graphicx}
\usepackage{subfigure}
\usepackage{booktabs}
\usepackage{amsmath}

\usepackage{amssymb}
\usepackage{mathtools}
\usepackage{amsthm}
\usepackage[capitalize,noabbrev]{cleveref}
\usepackage[textsize=tiny]{todonotes}
\usepackage[dvipsnames]{xcolor}

\newcommand{\modelname}{\texttt{TBS}}

\theoremstyle{plain}

\theoremstyle{definition}

\theoremstyle{remark}

\acmConference[SciSoc Agents \& LLMs '26]{KDD'26 Workshop on SciSoc Agents \& LLMs: Agentic AI for Scientific and Societal Advances}{August 9, 2026}{Jeju, Republic of Korea}
\acmYear{2026}
\copyrightyear{2026}

\title{Think-Before-Speak: From Internal Evaluation to Public Expression in Multi-Agent Social Simulation}

\author{Kaiqi Yang}
\email{kqyang@msu.edu}
\affiliation{
  \institution{Michigan State University}
  \city{East Lansing}
  \state{MI}
  \country{USA}
}

\author{Tai-Quan Peng}
\email{pengtaiq@msu.edu}
\affiliation{
  \institution{Michigan State University}
  \city{East Lansing}
  \state{MI}
  \country{USA}
}

\author{Sanguk Lee}
\email{lswook555@gmail.com}
\affiliation{
  \institution{Hankuk University of Foreign Studies}
  \city{Seoul}
  \country{Republic of Korea}
}

\author{Hui Liu}
\email{liuhui7@msu.edu}
\affiliation{
  \institution{Michigan State University}
  \city{East Lansing}
  \state{MI}
  \country{USA}
}

\renewcommand\footnotetextcopyrightpermission[1]{}
\begin{document}

\begin{abstract}
LLM-based multi-agent simulation offers a promising way to study social interaction, deliberation, and collective opinion dynamics. However, many existing dialogue simulation frameworks represent interaction mainly as observable turn exchange or aggregated outputs, leaving the internal evaluative processes behind silence, speaking intention, and public expression difficult to examine. We introduce \modelname{} (\textbf{T}hink-\textbf{B}efore-\textbf{S}peak), an interval-based multi-agent simulation framework that separates agents' private reasoning from public utterance generation. At each interval, all agents update structured internal states based on the shared dialogue history and their own memory. These states include dissonance-related appraisal, perceived opinion climate, perceived isolation risk, response strategy, and willingness to speak. The orchestrator then resolves competing speaking intentions and commits one utterance to the public dialogue, allowing internal evaluation and public interaction to co-evolve over time.

We evaluate \modelname{} in simulated town hall discussions on a climate-related policy issue. Results show that \modelname{} produces coherent internal-state traces and that these traces vary systematically across turn-allocation, silence, and memory conditions. Dissonance-related appraisal increases agents' willingness to speak, whereas silence-pressure appraisal decreases it. Once speaking intention is formed, public expression is shaped mainly by turn-allocation rules. These findings suggest that \modelname{} supports mechanism-sensitive social simulation by making the pathway from internal evaluation to public expression observable and analyzable.

\end{abstract}

\maketitle

\section{Introduction~\label{introduction}}

Recent advances in large language models (LLMs) have made it possible to simulate increasingly complex forms of social interaction, including deliberation, collective decision-making, and public discussion~\cite{Li2024ASO,Dubois2023AlpacaFarmAS}. These simulations offer a promising way to study how opinions, strategies, and participation patterns unfold in settings that are difficult to observe or experimentally manipulate at scale~\cite{gurcan2024llm,piao2025agentsociety,mou2026individual}. Yet most existing multi-agent dialogue frameworks still treat interaction primarily as a sequence of visible outputs. Agents are prompted to speak, their utterances are added to a shared transcript, and subsequent responses are generated from that transcript~\cite{Jeong2025SpeakTS,Zhang2025LLMAIDSimLA}. This design captures what is publicly said, but it provides only limited access to the internal evaluative processes through which participants interpret prior remarks, revise their positions, experience pressure, and decide whether speaking is warranted.

This limitation is especially important for simulating town hall discussions~\cite{WebbHooper2019UnderstandingMF,Etzioni1972MinervaAE}. In public deliberation, silence does not imply cognitive inactivity. Participants who are not speaking may still be listening, evaluating disagreement, reassessing the opinion climate, and deciding whether the current moment is safe or useful for expression~\cite{Taylor1982PluralisticIA,Maor2013OrganizationalRR}. Existing paradigms are poorly suited to this process. Hierarchical aggregation allows agents to reason in parallel, but it can create information asymmetry because agents do not fully respond to others' most recent contributions within the same round. Sequential turn-taking preserves a coherent public transcript, but it calls agents to reason mainly when their turn arrives, leaving the internal updates of non-speaking agents unmodeled~\cite{Li2024ASO,Feng2025GroupinGroupPO}. As a result, these approaches risk treating public utterances as direct products of the simulation protocol rather than as outcomes of an evolving internal process. The central challenge, therefore, is not only how to approximate continuous-time interaction with discrete computational steps, but also how to represent the pathway from internal evaluation to speaking intention and public expression. 
This challenge also connects to a broader shift in LLM-agent design from model-centric prompting to harness-oriented agent construction. Recent work on agent harnesses suggests that agent behavior is shaped not only by the backbone model, but also by the surrounding scaffold that organizes memory, protocols, observation, action mediation, feedback, and execution control~\cite{zhou2026externalization,xu2026runtimeharness}. Modular harness designs further show that perception, memory, and reasoning components can be composed around the model to support multi-turn agent behavior across interactive environments~\cite{zhang2025modularharness}. From this perspective, simulating deliberative interaction requires not only prompting agents to speak, but also designing a protocol that determines what agents remember, when they update internal states, how speaking opportunities are allocated, and which parts of the process become observable for analysis.

\paragraph{Research Questions.}
To address these challenges, we ask three research questions that connect the framework's architecture to its theoretical and empirical goals. Appendix~\ref{app:rq} provides fuller theoretical elaboration for each question.

\begin{itemize}
    \item[Q1] \textbf{How does interval-level internal-state updating affect the interpretability, analytical usefulness, and protocol-level cost structure of LLM-based social simulation?}
    This question focuses on whether the simulation can expose the latent evaluative processes that occur before public expression. In town hall discussion, non-speaking participants may still interpret prior remarks, update beliefs, and reassess their readiness to speak; TBS therefore examines whether interval-level updates can make these processes observable as structured traces rather than leaving them implicit in final utterances.

    \item[Q2] \textbf{Can LLM-based agents generate willingness to speak and public utterances as outcomes of ongoing internal evaluation, rather than as direct products of fixed speaking order or immediate reactive response?}
    This question distinguishes speaking intention from realized public speech. Instead of treating every agent response as an automatic consequence of a turn-taking protocol, we examine whether willingness to speak can emerge from agents' evolving internal states and whether public expression reflects both internal motivation and access to the floor.

    \item[Q3] \textbf{How do turn-allocation rules, silence constraints, and memory mechanisms influence communication strategies and opinion dynamics in a town hall setting?}
    This question connects the framework to controllable simulation design. By varying whether speaking is self-selected or externally assigned, whether silence is allowed, and how prior discussion is remembered, we analyze how temporal, social, and mnemonic constraints shape participation, response strategies, and opinion dynamics.
\end{itemize}

In this work, we propose \modelname, \textbf{T}hink-\textbf{B}efore-\textbf{S}peak, a discrete-time multi-agent framework that approximates continuous interaction through fine-grained temporal intervals. All agents update their reasoning at every interval based on both shared dialogue history and their own evolving internal states. Agents may independently attempt to speak, accompanied by an estimated response latency. The system interprets simultaneous speaking attempts as conflicts and resolves them by selecting the earliest responder, committing only one utterance per interval. This design distinguishes reasoning from speaking, allowing agents to update their internal states across intervals while ensuring a coherent, globally shared dialogue.

This framework offers several advantages. First, it provides a process-sensitive approximation of continuous interaction by enabling all agents to update their internal states at every step, rather than only when selected to speak. Second, it is designed to reduce redundant full-response generation by allowing agents to reuse and refine intermediate reasoning before public utterance generation. Third, by explicitly maintaining structured representations of internal states, strategies, and belief evolution, the framework enhances interpretability and supports fine-grained analysis of social behavior and decision processes~\cite{Park2024LLMAG}.

We summarize our contributions as follows:
\begin{itemize}
    \item We introduce a discrete-time simulation protocol that bridges continuous-time reasoning and discrete communication through interval-based interaction and conflict resolution.
     \item We propose a unified protocol that integrates parallel reasoning with competitive turn-taking, supporting a process-sensitive representation of deliberative interaction while making access to public expression explicit.
    \item We provide a structured and interpretable framework for tracking agent cognition and interaction dynamics, enabling downstream analysis in multi-agent systems and social science research.
    \item We present preliminary experiments showing that \modelname{} produces coherent internal-state traces and that dissonance-related appraisal, silence-pressure appraisal, turn-allocation, and memory conditions systematically shape speaking intention and public expression.
\end{itemize}
\section{Background and Related Work}
\subsection{Multi-Agent Social Simulation for Dialogue and Reasoning}

Recent advances in LLMs have enabled new forms of multi-agent dialogue, cooperation, and social simulation. Frameworks such as AutoGen~\cite{Wu2023AutoGenEN} and OASIS~\cite{yang2024oasisopenagentsocial} provide general infrastructures for composing multiple LLM agents, coordinating their conversations, and supporting autonomous task completion or role-playing interaction. These systems demonstrate the promise of LLM agents as interactive social entities rather than isolated text generators. Broader surveys of LLM-empowered agent-based simulation also suggest that LLMs can enrich traditional agent-based modeling by introducing more flexible reasoning, communication, and heterogeneous behavior~\cite{gurcan2024llm,piao2025agentsociety,mou2026individual}. However, existing frameworks are often highly integrated and task-oriented. Although some are open-sourced, their implementation may be difficult to modify when researchers need to introduce new theory-driven modules, especially for social scientists.

Another underexplored challenge is the dimension of time. Traditional agent-based simulation often relies on discrete time steps, even though real-world time is continuous and can be modeled as divisible into finer-grained events~\cite{Liu2024SpatialTemporalLL,Yu2023TemporalDM}. This gap becomes especially salient in LLM-based dialogue simulation, where agents must generate content within predefined turns or rounds. In open discussion, however, multiple people may attempt to speak at nearly the same time; moreover, every new utterance, and even the passage of time without speech, can change participants' internal reasoning. To address this limitation, we design a time-aware framework that separates thinking from speaking while making the additional reasoning steps explicit and controllable. Continuous interaction is abstracted into connected intervals, and a \texttt{time\_cost} component is introduced to coordinate cases where multiple agents simultaneously intend to speak.
Under the minimal constraint that speakers cannot be interrupted by others, the framework records all agents' internal trajectories throughout the discussion, rather than compressing multi-step reasoning into a single response only when an agent's public turn arrives.

\subsection{Harness Engineering for LLM Agents}

Recent work on LLM agents increasingly studies the harness as a key part of agent design. In this context, a harness refers to the surrounding scaffold that organizes how an agent receives observations, stores memory, follows protocols, invokes tools or actions, receives feedback, and records trajectories. A recent review summarizes harness engineering together with memory, skills, and protocols as important components for building reliable LLM agents~\cite{zhou2026externalization}. In task-oriented settings, harnesses have been used to evaluate LLM-guided programming assistants across code generation, documentation, testing, bug fixing, and workspace understanding~\cite{agarwal2024copilotharness}, and modular harnesses have been proposed for multi-turn gaming agents by composing perception, memory, and reasoning components~\cite{zhang2025modularharness}.

Another line of work studies how harnesses can be synthesized, governed, or repaired. AutoHarness automatically synthesizes code harnesses to prevent invalid actions in game environments~\cite{lou2026autoharness}. Harness-MU focuses on safe and governed multi-user LLM agents, where the harness coordinates permission control and compliant response generation~\cite{fan2026harnessmu}. HarnessFix diagnoses failures from agent trajectories and repairs harness flaws through trace-guided patching~\cite{chen2026harnessfix}. Recent analysis also shows that updating a harness and benefiting from the updated harness are distinct capabilities in self-evolving LLM agents~\cite{lin2026harnessupdating}.

Our work is complementary to these harness-oriented studies. Existing harness work mainly focuses on task success, valid action execution, safety governance, or failure repair. \modelname{} does not introduce a separately named harness component; instead, it incorporates harness-level design into the agent schema and orchestrator. The framework specifies how agents store, retrieve, and compress memory, how information is exchanged between agents and the shared discussion environment, and how private internal updates, intentions, and public utterances are recorded and broadcast under different rules. These design choices allow \modelname{} to coordinate agent behavior and preserve analyzable process traces within the simulation protocol. In this sense, \modelname{} adapts harness engineering to social simulation: the goal is not primarily to improve task completion, but to make the pathway from internal evaluation to speaking intention and public expression observable and analyzable in simulated deliberation.

\subsection{Opinion Dynamics and Response Strategies}

Town hall discussions involve not only public utterances, but also participants' ongoing interpretation of those utterances and their implications for later participation. This setting connects to research on cross-cutting exposure, political disagreement, and deliberative communication, which shows that opposing views can promote reflection while also creating participation costs for some citizens \citep{mutz2001facilitating, mutz2006hearing, eveland2009political}. Two social-psychological mechanisms are especially relevant. Cognitive dissonance \citep{festinger1957cognitive, metzger2020cognitive} suggests that conflicting statements may trigger tension and motivate defense, reinterpretation, revision, or withdrawal. Spiral of silence \citep{noelle1974spiral, matthes2018spiral} suggests that willingness to speak depends on perceived opinion support and isolation risk, not merely private opinion. Because perceived disagreement and actual disagreement can have distinct implications for deliberative experience \citep{wojcieszak2012perceived}, simulations should model participants' perceived inconsistency, perceived opinion support, and perceived isolation risk rather than only the objective distribution of opinions.

These mechanisms also show that opinion dynamics are inseparable from communication strategies. Participants may defend, qualify, partially align, remain silent, or wait for a more favorable moment to speak, reflecting research on willingness to self-censor \citep{hayes2005validating, hayes2005willingness}. In town hall settings, this process is especially important because all participants monitor the discussion, while only one speaker is publicly heard in a given turn. This combination of public visibility, perceived opinion climate, and constrained access to expression is central to contemporary spiral-of-silence research in mediated and networked settings \citep{neubaum2017opinion, chen2018spiral, gearhart2015was}. A simulation that only models observable turn exchange would miss how non-speaking participants update beliefs, reassess prior remarks, and revise willingness to speak. \modelname{} addresses this limitation by making the latent reasoning process underlying public expression analytically visible: agents continuously update internal states from shared dialogue history and evolving evaluations, while public speech remains temporally constrained and globally coherent.

Recent work on generative AI and deliberation has developed along two lines. One treats AI as an intervention in human discussion: AI-generated rephrasing can improve political conversations by encouraging listening, validation, and respect \citep{argyleLeveragingAIDemocratic2023}, while the Habermas Machine synthesizes opinions and critiques into group statements that capture common ground \citep{tesslerAICanHelp2024}. A second line uses AI for deliberative simulation. Generative AI has been proposed as a deliberation-making tool for training, policy consultation, classroom deliberation, and theory development \citep{rountreeCaseUsingGenerative2026}, and systems such as Plurals use persona-based LLM agents and moderators to simulate diverse perspectives under different interaction rules \citep{ashkinazePluralsSystemGuiding2025}. However, existing approaches tend to emphasize improving human deliberation or generating deliberative outputs, while leaving undertheorized how participation emerges during an unfolding public discussion. Our study addresses this gap by modeling deliberation as an interval-level communication process in which agents listen, update internal states, remain silent, and decide whether to speak. Conceptually, this shifts attention from AI-assisted deliberation and deliberative output generation to the simulation of deliberative communication; methodologically, it provides a mechanism-sensitive architecture for studying latent communicative processes that are difficult to observe directly.
\section{Notation and Preliminaries}

Existing multi-agent simulation paradigms can be broadly categorized into hierarchical aggregation and sequential turn-taking. While both provide workable abstractions for coordinating multiple agents, they introduce structural limitations in information flow and reasoning continuity, especially when approximating continuous-time interaction with discrete computational steps.

\subsection{Notation}
Let $\mathcal{R}=\{r_1, r_2, ... , r_{|\mathcal{R}|}\}$ denote the set of interaction rounds $r$; in each round $r_i$, we further introduce the set of intervals $\mathcal{V}_i=\{v_1^i, v_2^i, ... , v_{|\mathcal{V}_i|}^i\}$, in which each $v$ corresponds to a single speaking opportunity. We use \textbf{$\prec$} to denote the temporal sequence, e.g., $r_1 \prec r_2$.

Let $\mathcal{A} = \{a_1, a_2, ..., a_{|\mathcal{A}|}\}$ denote the set of agents $a$. Generally, we use $X$ to denote the text information from aggregation or LLMs' generation. Specifically, the agents can potentially generate internal states $s$ and utterances $u$. Here, $s$ is the internal state containing reasoning, perception of dialogue context, and response strategy, while $u$ is the utterance (spoken content) that is captured. Besides, $Mem$ stands for agents' memory, including the memory of group dialogue, the agent's own inner states, and the speaking actions.

For convenience in describing the text generation and concatenation of LLMs, we denote by $\Theta(\cdot)$ the text generation function of agents (e.g., an LLM inference step) with certain prompts. We define the concatenation operator as below, which concatenates a sequence of text segments in order.
\[
\Psi_{i=1}^{n} text_i \;:=\; text_1 \,\|\, text_2 \,\|\, \cdots \,\|\, text_n.
\]

\subsection{Baseline Frameworks}
\label{sec:baseline_frameworks}

We describe the two baseline paradigms below; detailed schematic illustrations are provided in Appendix~\ref{app:baseline_figures}.

\paragraph{Hierarchical Aggregation}
In this framework, there is only one single interval $v^i_1$ in each round\footnote{In this setting, interval $v$ degrades to the equivalence of round $r$; for simplicity, we use $v^i_1$ and $v^i$ interchangeably only in this setting.}, and all agents generate outputs in parallel based on the previous rounds. Formally, for agent $\alpha$ in round $\tau$, with the set of prior rounds $\mathcal{R}=\{r\ |\ r\prec\tau\}$:
\begin{equation}
\begin{aligned}
X_{GroupContext} = &\ \bigl(\Psi_{a \in \mathcal{A}}\ \Psi_{r\in\mathcal{R}}\ u(r,a)\bigl) \\
X_{SelfThought} = &\ \bigl(\Psi_{r\in\mathcal{R}}\ s(r,\alpha)\bigl) \\
\{s(\tau,\alpha), u(\tau,\alpha)\} = & \ \Theta\bigl(X_{GroupContext},\ X_{SelfThought}\bigl)\\
\end{aligned}
\end{equation}

A coordinator captures all utterances $u_\tau$, and then broadcasts the aggregated utterances or the concluded content. 

Although this design follows a socially hierarchical structure, it leads to information asymmetry. Specifically, agent $\alpha$ derives its reasoning $s(\tau,\alpha)$ and utterance $u(\tau,\alpha)$ without considering the outputs of other agents $a\prime$ within the same round $\tau$, i.e., $u(\tau,a\prime)$. In other words, both agents $\alpha$ and $a\prime$ base their decisions solely on the dialogue context from rounds $r \prec  \tau$. As a result, $u(\tau,a\prime)$ may either fail to address or redundantly repeat points that have already been expressed in $u(\tau,\alpha)$.

\paragraph{Sequential Turn-taking}
In sequential turn-taking, each round $r_i$ consists of $\mathcal{A}_i$ intervals, as each agent speaks per interval. For the selected agent $\alpha$, the generation is given by:
\begin{equation}
\begin{aligned}
X_{GroupContext} = &\ \bigl(\Psi_{v\in\mathcal{V}}\ u(v)\bigl) \\
X_{SelfThought} = &\ \bigl(\Psi_{r\in\mathcal{R}\backslash\tau}\ s(r,\alpha)\bigl) \\
\{s(\tau,\alpha), u(\tau,\alpha)\} = & \ \Theta\bigl(X_{GroupContext},\ X_{SelfThought}\bigl)\\
\end{aligned}
\end{equation}

while other agents remain inactive. Please note that the set of current intervals $\mathcal{V}$ contains both all intervals in prior rounds and the intervals before calling agent $\alpha$; as each interval has its sole speaker, we omit the notation $a$ in utterance $u$. Although this ensures consistent dialogue history, it calls agents to think and speak only when it is their turn. This design neglects intermediate reasoning updates, including being persuaded, refining others' opinions, and feeling pressured by one's utterances, which are of vital importance in social and behavioral studies. 

Besides, this framework can create a protocol-level trade-off between agent-calling cost and participation opportunities. Denote $T$ as the cost of agent calling (i.e., count of intervals), and $t$ as the chances each agent has to think and speak. As $\mathcal{A}$ intervals are required per round, this leads to either higher overhead if the number of rounds $\mathcal{R}$ is fixed ($T = |\mathcal{A}| \times |\mathcal{R}|$) or fewer participation chances for each agent ($t=\lfloor \hat{t}\rfloor$ or $\lceil \hat{t}\rceil$ with $\hat{t}=T/|\mathcal{A}|$) when the budget $T$ is fixed.


\section{\modelname{}: Time-Aware Social Simulation}

In this section, we introduce \modelname, a flexible multi-agent framework that manages agents' speaking and thinking through a controllable interaction pipeline. The framework supports interval-based interaction, continuous internal reasoning, and conflict-resolved speaking allocation. We first describe the agent design, including agents' core abilities such as reasoning, memory management, and speaking-decision making. We then introduce the orchestrator and explain how it coordinates agents and guides the dialogue process.

\subsection{Agent Schema}

To support both the thinking and speaking processes, each agent is equipped with an internal reasoning module and speaking module. The first module underpins how the agent interprets the ongoing discussion and updates its internal state. This module captures multiple aspects of social cognition, including the perception of others' utterances, the evaluation of agreement or conflict with the agent's own position, and the estimation of psychological and social pressures during the interaction. Specifically, the internal state $s$ contains the following fields:
\begin{itemize}
    \item \texttt{internal\_state}: a natural-language description of the agent's internal reasoning process. This field records how the agent perceives, interprets, and evaluates the ongoing dialogue, together with the reasoning process that leads to subsequent actions.

    \item \texttt{time\_cost}: the estimated time (in seconds) required for the agent to complete the reasoning process and decide whether and how to respond. This field is introduced to simulate time-sensitive discussion scenarios in which only one participant can speak at a given moment, although multiple agents may simultaneously reason about the dialogue and express willingness to speak. 

    \item \texttt{current\_opinion}: the agent's current stance on the discussion topic in a concise natural-language form.

    \item \texttt{perceived\_inconsistency}: the degree (0 to 1) to which the statements made by other participants are perceived as conflicting with the agent's current opinion.

    \item \texttt{dissonance\_tension}: the degree (0 to 1) of internal psychological tension induced by the perceived inconsistency between external opinions and the agent's own stance.

    \item \texttt{motivation\_to\_reduce\_dissonance}: the extent (0 to 1) to which the agent is motivated to reduce such tension, for example through defending, reinterpreting, or adjusting its position.

    \item \texttt{perceived\_opinion\_climate}: the agent's perception of the overall opinion climate within the discussion, ranging from strongly opposing (indicated by -1) to strongly supporting (indicated by 1) the agent's own view.

    \item \texttt{perceived\_isolation\_risk}: the perceived level of social risk (0 to 1) associated with publicly expressing the agent's opinion in the current group setting.

    \item \texttt{response\_strategy}: the conversational strategy selected by the agent for the current interaction step. The agent selects one strategy from the strategy pool: \texttt{defend, qualify, converge, challenge, bridge, withhold}\footnote{Here, \texttt{withhold} indicates that the agent chooses not to speak in the current round based on the dialogue context and its internal states.}.
\end{itemize}

These fields are theory-guided rather than purely ad hoc: the inconsistency, tension, and motivation fields operationalize dissonance-related appraisal, whereas perceived opinion climate and isolation risk operationalize silence-pressure appraisal derived from spiral-of-silence and self-censorship research \citep{festinger1957cognitive, hayes2005validating, hayes2005willingness, matthes2018spiral, metzger2020cognitive, noelle1974spiral}.

The second module, the speaking module, generates the utterance for the agent selected to speak. The generated text is recorded as the new public dialogue content and broadcast to the shared environment, where it serves as input for agents' reasoning in subsequent intervals. Under this setting, each dialogue round $r$ contains exactly one speaking interval $v$, while all agents remain cognitively active through internal reasoning.

\subsection{Memory Management}

To help agents track the evolving dialogue while reducing input-token consumption, we design a three-part memory schema, $Mem$, for each agent: group memory, monologue memory, and self memory. The memory module serves as a structured database that stores: (1) \textit{group memory}, which records public utterances $u(r)$ and their corresponding speakers $a(r)$; (2) \textit{monologue memory}, which stores the agent's own internal thinking states $s$ from previous rounds; and (3) \textit{self memory}, which records the agent's own response strategies and spoken content when it is allowed to speak. Importantly, only public utterances and speaker names are visible to other agents, whereas internal monologues remain private and are accessible only to the corresponding agent.

In addition, we introduce dynamic memory retrieval to approximate selected aspects of forgetting and selective recall. Each type of memory can be stored and retrieved through one of three mechanisms: (1) \texttt{latest}, where only the most recent $n$ rounds are retained; (2) \texttt{random}, where previous rounds are randomly sampled according to a long-tail distribution that assigns higher sampling weights to more recent memories, reflecting the assumption that newer information is less likely to be forgotten; and (3) \texttt{summary}, where agents compress prior dialogue and internal states into summaries. This summary-based mechanism approximates memory accumulation, loss of early details, and selective retention shaped by the agent's stance and preferences.

\subsection{Orchestrator}

Given the agents described above, the proposed pipeline maintains an orchestrator as the management server that coordinates both the agents and the discussion process. The orchestrator is not a human-like agent with a specific profile; instead, it operates through objective rules that control turn allocation, information flow, and memory updates. At each orchestrator step, referred to as an interval $v$, the following operations are executed in order:

\begin{itemize}
    \item \texttt{Encourage Thinking}: The orchestrator asks all agents to reason over the prior dialogue, together with their three types of memory, and to formulate their current response intentions.
    
    \item \texttt{Speaking Chance Allocation}: The orchestrator collects each agent's response intention and selects one speaker for the current interval $v$. As defined by \texttt{time\_cost}, when multiple agents intend to respond to the same prior utterance, the agent with the smallest estimated time cost, i.e., the fastest reasoning process, is selected as the next speaker, while the speaking intentions of the other agents are temporarily withheld.
    
    \item \texttt{Utterance Broadcast}: After the selected agent produces an utterance $u$, the orchestrator broadcasts it to all agents. This allows each agent to incorporate the new public utterance into its internal state and continue reasoning in the next interval.
\end{itemize}

This design separates agents' internal reasoning from their spoken actions and makes the intermediate process available for analysis. At each interval, agents reason over the available dialogue context, including prior responses within the same round, which enables fine-grained tracking of evolving internal states rather than relying only on observable speech. Meanwhile, non-speaking agents update their internal reasoning without generating full public utterances, so the framework is designed to avoid some redundant full-response generation. The speaking step then uses the selected agent's updated internal state as a reference. Overall, the orchestrator coordinates richer social information while preserving a coherent public dialogue and explicit records of how internal states, speaking intentions, and public expression are connected.

\begin{figure*}[ht]
\vskip -0.1in
\begin{center}
\centerline{\includegraphics[width=0.75\textwidth]{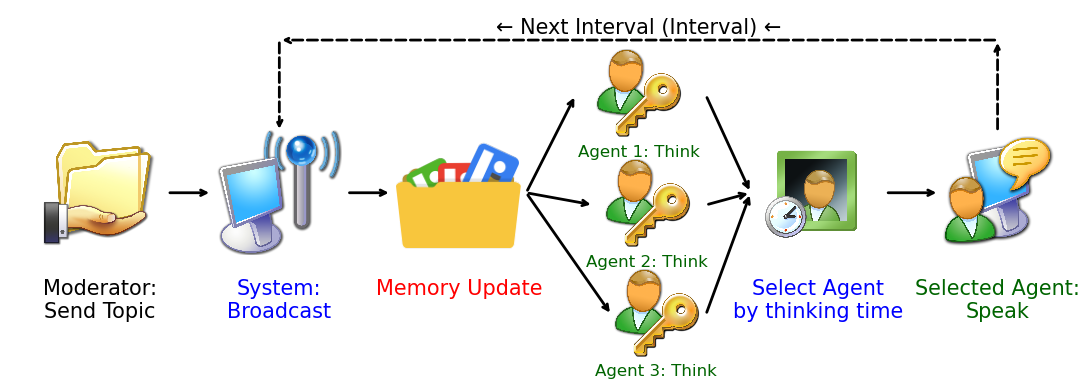}}
\caption{Framework of \modelname{} System.}
\label{TimeSimulator}
\end{center}
\vskip -0.1in
\end{figure*}

\section{Experiments}

To evaluate the framework, we run simulations on socially important topics with human-derived profiles and analyze the generated discussion logs from the perspective of dialogue and communication studies.

\subsection{Experimental Setup}
We use a town hall discussion as the task scenario and testbed for our simulation framework. The discussion is initialized with a curated topic and a set of human profiles. 
As an initial testbed, we focus on the climate-related topic of a ``solar photovoltaic (PV) mandate'' and construct two sets of agent personas: Six Americas and Balanced Stakeholders~\cite{leiserowitz2021global}, both of which provide representative taxonomies for climate-related attitudes.
The \texttt{Six Americas} personas cover six attitudes toward climate change: alarmed, concerned, cautious, disengaged, doubtful, and dismissive. The \texttt{Balanced Stakeholders} personas include three pairs of profiles, where each pair contains one supportive and one opposing stakeholder, and their other characteristics are equivalent or comparable. 

In terms of the backbone LLMs powering the agents, we use Gemini-2.5-Flash-Lite and Gemini-2.5-Flash in the Colab environment, without specifying additional system prompts or tuning parameters. 
Unless otherwise stated, the results reported in this paper are based on Gemini-2.5-Flash-Lite. We also use Gemini-2.5-Flash for qualitative robustness checks, but we leave systematic cross-backbone evaluation to future work.

\paragraph{Turning Mode} 
To control the allocation of dialogue intervals, following prior work, we deploy two turn-allocation modes. The \texttt{willing} mode maintains an open discussion setting, where agents autonomously apply for speaking opportunities. When multiple agents express willingness to speak within the same interval, the system selects the agent that applies first (i.e., having the shortest reasoning time) and suppresses the others. In contrast, the \texttt{rotation} mode manages a predefined speaking order, where agents are prompted to speak sequentially in each round.

\subsection{Interaction Schema}

We further examine whether agents are allowed to remain silent, a design choice closely related to spiral-of-silence and self-censorship theories of public opinion expression \citep{noelle1974spiral, hayes2005validating, hayes2005willingness, matthes2018spiral}. In each interval, agents first decide whether they are willing to speak. Under the \texttt{w/ Force Speak} setting, if no agent expresses willingness to speak, the orchestrator randomly selects one agent and requires it to produce an utterance. This setting ensures that the discussion continues even when no agent voluntarily takes the turn. In contrast, under the \texttt{w/o Force Speak} setting, agents may all remain silent when none of them is willing to speak. This preserves full autonomy in speaking decisions and allows silence to become an observable outcome of the interaction process.

\subsection{Memory Mechanism}
To approximate selected aspects of memory decay while controlling retrieved context length, we introduce memory modes that dynamically retrieve historical dialogue context to support agents' reasoning and speaking processes.

The \texttt{latest} mode adopts a hyperparameter $n_{Memory}$ as the maximum memory capacity. When the number of dialogue rounds exceeds $n_{Memory}$, only the most recent $n_{Memory}$ rounds are retained. This mode reflects a simple forgetting mechanism, assuming that only the most recent information remains salient.

The second mode, \texttt{random}, introduces a hyperparameter $n_{Random}$.
When the number of stored dialogue rounds exceeds this threshold, the system samples
$n_{Random}$ rounds according to a recency-weighted power-law distribution.
Specifically, for the currently available dialogue history
$\mathcal{R}_{cur}=\{r_1,r_2,\ldots,r_T\}$ with
$r_1 \prec r_2 \prec \cdots \prec r_T$, each round $r_j$ is assigned a weight $w(r_j)=\ j^{\ 1.3}$, which is then normalized as $p(r_j)=w(r_j)/\sum_{k=1}^{T}w(r_k)$.
The system samples rounds according to these normalized probabilities.
Thus, later rounds receive larger sampling probabilities, while earlier rounds still
have a non-zero chance of being retrieved. This design approximates the phenomenon
that more recent dialogue is more likely to be remembered, while still allowing
occasional retention of older information.

The third mode, \texttt{summary}, leverages LLMs to iteratively compress dialogue history. Given the summarization function as $\tilde{\Theta}(\cdot)$ and $Mem_0=(a(0), u_0)$, the following sequence of memory $\{Mem_k\ |\ k>1\}$ is generated by:
\[
Mem_k \;=\; \tilde{\Theta}\ \bigl(Mem_{k-1},\, a(k),\, u_k\bigl),
\]
where $a(k)$ denotes the agent that produces the utterance $u_k$ in the time interval $v_k$.

Intuitively, at dialogue interval $v_k$, the most dominant memory component is derived from the latest utterance $u_k$ and its speaker $a(k)$, while earlier dialogue contributes with progressively diminishing influence. Moreover, the summarization function $\tilde{\Theta}$ is conditioned on agent profiles, capturing the phenomenon that different agents may form distinct interpretations and memories even when observing the same utterances.

\section{Results}

We analyze whether~\modelname{} produces interpretable internal-state traces and whether these traces help explain speaking intention and public expression. Because the simulations generate open-ended dialogue, our analyses focus on structured process indicators extracted at the agent-interval level: dissonance-related appraisal, perceived silence pressure, willingness to speak, and public expression. The first two indicators summarize agents' interval-level internal evaluations, while the latter two represent the transition from internal evaluation to public communication. Full index construction details, model specifications, and additional results are reported in Appendix~\ref{app:result}.

\subsection{Internal-State Indices and Dynamics}

A central goal of \modelname{} is to make agents' internal evaluations observable as structured process traces. We constructed two theory-guided indices. The dissonance index averaged perceived inconsistency, dissonance tension, and motivation to reduce dissonance. These indicators correspond to whether an agent perceives the ongoing discussion as conflicting with its own position, whether that inconsistency produces tension, and whether the agent is motivated to resolve or reduce the tension \citep{festinger1957cognitive, metzger2020cognitive}. The silence-pressure index averaged unfavorable perceived opinion climate and perceived isolation risk. Since the original perceived opinion climate variable ranged from $-1$ to $1$, it was first reverse-coded and rescaled so that larger values represented a more unfavorable climate for the agent's own position. This transformation makes the direction of the index consistent with perceived isolation risk: higher values indicate stronger pressure against public expression \citep{noelle1974spiral, hayes2005willingness, matthes2018spiral}.

Both indices showed strong internal coherence. The dissonance index yielded Cronbach's $\alpha = .89$, 95\% CI [.88, .89], and an average inter-item correlation of .72. The silence-pressure index yielded Cronbach's $\alpha = .92$, 95\% CI [.92, .92], and an average inter-item correlation of .87. Because the silence-pressure index contains only two indicators, the inter-item correlation is especially important for interpretation. These results suggest that the framework produces internally consistent process indicators rather than isolated prompt outputs. These indices should be interpreted as structured process traces generated by the simulation, not as validated psychometric measures of human psychological states. Their purpose is to make agents' latent evaluations analyzable during the unfolding discussion.

We then estimated linear mixed-effects models to examine how these indices varied across interval and experimental conditions. Each model used one internal-state index as the outcome and included centered interval, persona ecology, turn-allocation rule, Force Speak setting, and memory mode as fixed effects. To test whether temporal trajectories differed by simulation design, we also included interactions between centered interval and each design factor. Random intercepts were included for both simulation run and agent, accounting for repeated observations within runs and within agents. This specification separates average condition differences from condition-specific changes over the course of the discussion.

The dissonance model showed that both design conditions and temporal trajectories shaped dissonance-related appraisal. Compared with turn-taking, the willing mode was associated with higher dissonance, $b = .130$, $p = .030$, and Force Speak was also associated with higher dissonance, $b = .089$, $p = .048$. The six-Americas persona ecology was associated with lower dissonance than the balanced-stakeholder persona ecology at the average interval, $b = -.132$, $p = .006$. Memory mode did not produce significant main effects on the average dissonance level.

The temporal coefficients indicate that dissonance increased modestly over the discussion, $b = .00043$, $p = .003$. This growth was not uniform across conditions. The increase was stronger in the six-Americas condition, $b = .00053$, $p < .001$, and much stronger in the willing mode, $b = .00886$, $p < .001$. The summary memory condition reduced the rate of increase, $b = -.00071$, $p < .001$. The interval-by-Force Speak interaction was not significant, suggesting that Force Speak affected the average level of dissonance but did not significantly change its temporal slope. These findings indicate that self-selected participation changes the internal dynamics of the discussion: agents experience greater contradiction and stronger tension when public expression depends on voluntary entry into the floor.

The silence-pressure model showed a different configuration of effects. The willing mode was associated with higher silence pressure than the turn mode, $b = .145$, $p = .006$. Force Speak was also associated with higher silence pressure, $b = .069$, $p = .033$. Memory mode showed a clear main effect for summary memory: compared with the \texttt{latest} mode, summary memory increased silence pressure, $b = .110$, $p = .008$, while \texttt{random} memory did not differ significantly from the \texttt{latest} mode. The six-Americas condition did not significantly differ from the balanced-stakeholder condition at the average interval, $b = -.006$, $p = .850$.

Unlike dissonance, silence pressure did not exhibit a significant average linear trend over time, $b = .000004$, $p = .977$. Yet its trajectory still differed across simulation settings. Silence pressure increased faster in the six-Americas condition, $b = .00089$, $p < .001$, in the willing mode, $b = .00670$, $p < .001$, and in the summary memory condition, $b = .00081$, $p < .001$. The interaction between interval and Force Speak was not significant. This pattern suggests that silence pressure is especially sensitive to how participation and memory shape the perceived opinion climate. Summary memory increased both the average level and the growth rate of silence pressure, possibly because summarized memory makes the emerging climate more salient by compressing prior remarks into a more accessible representation of the discussion's direction.

Together, these findings address RQ1 and RQ3 by showing that interval-level internal-state tracking produces coherent, analyzable process traces and that these traces vary systematically across turn-allocation, silence, and memory conditions. The divergence between the two indices is also important. Summary memory was associated with weaker growth in dissonance but stronger growth in silence pressure. This suggests that summarized memory may reduce the salience of discrete contradictions while increasing the perceived coherence of the overall opinion climate. Therefore, dissonance-related appraisal and silence-pressure appraisal should not be treated as interchangeable indicators of general negativity. The former centers on perceived inconsistency and tension, while the latter centers on social risk and opinion climate.

\subsection{Internal Evaluation, Speaking Intention, and Public Expression}

We next examined whether agents' internal evaluations predicted willingness to speak, directly addressing RQ2. A logistic mixed-effects model predicted whether an agent wanted to speak at each interval from the two internal-state indices, centered interval, persona ecology, turn-allocation rule, Force Speak setting, memory mode, and interactions between turn-allocation rule and each internal-state index. Random intercepts were included for run and agent. The internal-state indices were rescaled so that coefficients represented a .10-unit increase in the original index.

Internal evaluations strongly predicted speaking intention. Higher dissonance-related appraisal was associated with a higher probability of wanting to speak, $b = 1.47$, $p < .001$. On the rescaled metric, this means that a .10-unit increase in dissonance was associated with approximately 4.34 times greater odds of wanting to speak. In contrast, higher silence pressure was associated with lower willingness to speak, $b = -1.43$, $p < .001$. A .10-unit increase in silence pressure corresponded to an odds ratio of approximately 0.24. Thus, dissonance functioned as an expressive motivator, whereas silence pressure functioned as an expressive constraint.

These results align with the theoretical motivation of the framework. Dissonance-related appraisal acts as an expressive motivator: when agents perceive inconsistency, tension, or a need to reduce dissonance, they become more likely to enter the discussion. Silence pressure acts as an expressive constraint: when agents perceive the climate as unfavorable or socially risky, they become less likely to express themselves. This pattern is consistent with spiral-of-silence and self-censorship research on perceived minority status, unfavorable opinion climate, and isolation concerns \citep{noelle1974spiral, matthes2018spiral}.

The strength of these relationships differed by turn-allocation rule. Compared with turn-taking, the willing mode weakened the positive relationship between dissonance and willingness to speak, $b = -0.74$, $p < .001$, and weakened the negative relationship between silence pressure and willingness to speak, $b = 0.70$, $p < .001$. However, the estimated within-willing effects retained the same direction: dissonance remained positive, $b = 0.73$, and silence pressure remained negative, $b = -0.74$. This pattern suggests that self-selected participation does not remove the role of internal evaluation; rather, it changes how strongly internal states translate into speaking intention.

The main effects of other design factors provide additional context. Agents in the six-Americas persona ecology were less likely to want to speak than agents in the balanced-stakeholder condition, $b = -1.18$, $p < .001$. The willing mode also had a negative main effect on willingness to speak at average internal-state levels, $b = -1.83$, $p < .001$. This likely reflects the higher expressive threshold created when agents must self-initiate participation rather than respond to an externally assigned turn. Force Speak and memory mode did not show significant direct effects after internal-state indices and design factors were included.

Finally, we examined whether speaking intention became public expression. This analysis was restricted to agent-interval observations in which the agent had formed a speaking intention, and the outcome was whether that intention became the publicly selected utterance in the corresponding interval. Once agents already wanted to speak, internal evaluations were only weakly associated with whether they received the public floor. Dissonance-related appraisal was positively but marginally associated with being allowed to speak, $b = 0.08$, $p = .054$. Silence-pressure appraisal was negatively but marginally associated with being allowed to speak, $b = -0.07$, $p = .082$. These coefficients are small compared with the corresponding effects in the speaking-intention model, showing that internal evaluations are more important for forming willingness than for determining final access to public expression.

By contrast, turn-allocation rule strongly predicted realized expression. Among agents who wanted to speak, the willing mode increased the odds of being allowed to speak, $b = 1.21$, $p < .001$, corresponding to approximately 3.34 times greater odds. This is consistent with the design of the willing mode: agents autonomously apply for speaking opportunities, and the system selects among those who seek the floor. The interaction results gave only limited evidence that internal-state effects on public expression differed by turn-allocation rule. The dissonance-by-willing interaction was negative and marginally significant, $b = -0.21$, $p = .093$, while the silence-pressure-by-willing interaction was not significant, $b = -0.08$, $p = .531$.

Taken together, these models suggest a two-stage process. In the first stage, internal evaluations primarily shape whether agents develop a speaking intention: dissonance motivates expression, while silence pressure constrains it. Once an intention exists, public expression is shaped mainly by the rules governing access to the floor. This distinction supports the central premise of~\modelname: observable speech is not equivalent to internal willingness, but emerges from the combination of internal communicative motivation and external turn-allocation structure.
\section{Conclusion and Discussion}

Our framework provides a theory-guided and process-oriented testbed for open-ended social simulation by separating agents' private reasoning from public utterances. Through interval-based interaction, continuous memory updates, and willing-mode participation, agents first evaluate the evolving discussion, decide whether to speak, and then compete for limited speaking opportunities under rules of the orchestrator. This design represents theory-relevant social processes such as cognitive dissonance, group pressure, and spiral-of-silence dynamics, while making them observable as fine-grained trajectories of internal states, speaking intentions, and public expression. Thus, the framework supports process-level, interpretable, and experimentally controllable social simulation grounded in established social science theories.

This work remains preliminary and has several limitations. First, the internal states are generated by LLM agents under a theory-guided schema and should not be interpreted as validated measurements of human psychological states. Although the traces are internally coherent and analytically useful within the simulation, future work should validate them against human behavioral data, external annotations, or expert judgments of deliberative interaction. Second, memory management relies on heuristic retrieval strategies, including retaining the latest $n$ records, probability-based sampling, and iterative summarization. Although these approximate temporal memory decay, they remain coarse and are not grounded in psychological theories of memory. Third, we do not use RAG or similarity-based retrieval because human recall is not purely optimized for task relevance, but memories can also be triggered by contextual similarity. Future work should balance temporal decay with content-based reactivation. Fourth, due to computational cost, our experiments cover limited topics, language models, and agent scales, restricting cross-domain comparison and analysis of emergent effects as agent numbers increase. Future studies will expand the experimental scale to investigate questions relevant to both social science and computer science.

\clearpage
\bibliographystyle{ACM-Reference-Format}
\bibliography{ref}

\appendix
\appendix

\section{Theoretical Elaboration on Research Questions}
\label{app:rq}

This section provides a fuller explanation of the research questions guiding the study. The three questions are designed to connect the theoretical motivation of the framework with its simulation architecture and empirical analyses. Specifically, they ask how interval-level internal-state updating changes the observability and cost structure of LLM-based social simulation, whether agents' willingness to speak and public utterances can emerge from ongoing internal evaluation, and how design factors such as turn-allocation rules, silence constraints, and memory mechanisms shape communication strategies and opinion dynamics.

\textbf{Q1: How does interval-level internal-state updating affect the interpretability, analytical usefulness, and protocol-level cost structure of LLM-based social simulation?}

Many existing LLM-based social simulations represent interaction primarily through observable turn exchange. In such settings, agents are often modeled as producing visible responses when prompted, while the internal evaluative processes that precede public expression remain implicit or unmodeled. This creates a limitation for studying communication processes, because communication theory suggests that discussion does not unfold only through what participants say. It also depends on how participants interpret prior remarks, update their beliefs, reassess the relevance of the issue, and decide whether they are ready or willing to speak.

This distinction is especially important in a town hall discussion, where participants may continue listening, evaluating, and revising their internal positions even when they do not publicly speak. Silence, therefore, does not imply inactivity. Instead, latent processes remain active between public turns and may shape later participation. Building on the formulation in the main text, Q1 asks whether interval-level internal-state updating can make these latent processes more observable and analytically tractable, while also changing how computational effort is distributed between private state updates and public utterance generation.

This question is directly suited to \modelname, which maintains structured traces of evolving internal states and allows agents to reuse and refine intermediate reasoning across intervals. By making these traces observable, the framework allows researchers to analyze not only final utterances but also the intermediate process through which agents interpret the discussion, form speaking intentions, and eventually produce or withhold public expression. The cost-related part of the question is therefore treated as a protocol-level trade-off rather than as a claim of demonstrated efficiency improvement.

\textbf{Q2: Can LLM-based agents generate willingness to speak and public utterances as outcomes of ongoing internal evaluation, rather than as direct products of fixed speaking order or immediate reactive response?}
    
A central implication of communication theory is that public expression is not equivalent to private cognition. Participants do not automatically speak whenever they have an opinion. Instead, they first interpret what others have said, evaluate whether those remarks conflict with or support their own position, assess whether the current moment is appropriate, and then decide whether speaking is warranted. This distinction matters for theories of selective participation, where silence, hesitation, and delayed response may carry substantive meaning rather than merely indicating an absence of generated content.

We therefore ask whether LLM-based agents can generate willingness to speak and public utterances as outcomes of ongoing internal evaluation. This question contrasts such a mechanism with simpler approaches in which public speech is produced directly from fixed speaking order or immediate reactive response. It is directly supported by our simulation protocol, which compares externally assigned turn-taking with a willing mode. In the willing mode, speaking opportunities are endogenously claimed on the basis of evolving internal states. This allows us to examine whether agents' decisions to speak are linked to internal evaluations rather than only to procedural turn assignment.

\textbf{Q3: How do turn-allocation rules, silence constraints, and memory mechanisms influence communication strategies and opinion dynamics in a town hall setting?}

Communication in a town hall is shaped not only by participants' internal evaluations, but also by the social and temporal conditions under which expression occurs. Participants may behave differently depending on whether expression is self-selected or externally imposed, whether silence remains a meaningful option, and how prior discussion is remembered and incorporated into subsequent judgment. These conditions influence not only who speaks, but also how agents interpret the discussion, manage disagreement, revise their readiness to speak, and select communication strategies.

We therefore ask how turn-allocation rules, silence constraints, and memory mechanisms influence agents' participation and expression in the simulated discussion. This question is supported by the design of the framework, which varies each of these components explicitly. By doing so, the study can analyze how temporal constraints, social constraints, and mnemonic constraints shape communication strategies and opinion dynamics over time. This allows the simulation to examine not only the content of public utterances, but also the conditions under which willingness, silence, and expression emerge.

\section{Additional Background on Opinion Dynamics, Response Strategies, and AI-Mediated Deliberation}

In a town hall discussion, opinion dynamics unfold not simply through the sequence of public utterances, but through participants' ongoing interpretation of those utterances and their consequences for subsequent participation. This setting is closely related to research on cross-cutting exposure, political disagreement, and deliberative communication, which shows that encountering opposing views can promote reflection while also creating participation costs for some citizens \citep{mutz2001facilitating, mutz2006hearing, eveland2009political}.

This perspective is consistent with a long tradition in social science that treats communication behavior as mediated by latent psychological and social appraisals. In the present context, two mechanisms are especially relevant. Cognitive dissonance \citep{festinger1957cognitive, metzger2020cognitive} suggests that when participants encounter statements that conflict with their existing comments, they may experience tension that motivates defense, reinterpretation, revision, or withdrawal. Spiral of silence \citep{noelle1974spiral, matthes2018spiral} suggests that willingness to speak depends not only on private opinion, but also on whether that opinion appears socially supported within the interaction. Participants may therefore remain silent even when they hold strong views, particularly when the emerging opinion climate appears unfavorable.

Importantly, these processes are likely to depend less on the objective distribution of opinions than on participants' perceptions of the discussion climate. Prior research on deliberation shows that perceived disagreement and actual disagreement can have distinct implications for deliberative experience \citep{wojcieszak2012perceived}. This distinction is especially relevant for simulation, because agents may update their willingness to speak based on perceived inconsistency, perceived opinion support, and perceived isolation risk rather than on the full underlying distribution of positions in the group.

Taken together, these mechanisms highlight that opinion dynamics are inseparable from communication strategies. Participants do not merely hold opinions; they also decide how those opinions will be managed in public interaction. A participant may openly defend a position, qualify it, align partially with others, remain strategically silent, or wait for a more favorable moment to contribute, reflecting broader research on willingness to self-censor in public opinion expression \citep{hayes2005validating, hayes2005willingness}. These choices are shaped by ongoing internal evaluations of belief compatibility, social support, and expressive risk. In this sense, communication strategy is not external to opinion dynamics, but one of the principal ways through which opinion dynamics become visible in discussion.

The town hall setting makes these internal processes especially consequential. Although all participants continuously monitor the discussion, only one speaker is publicly heard in a given turn. This combination of public visibility, perceived opinion climate, and constrained access to expression is central to contemporary extensions of spiral-of-silence research in mediated and networked settings \citep{neubaum2017opinion, chen2018spiral, gearhart2015was}. Non-speaking participants are not passive. They continue to update their beliefs, reassess the implications of prior remarks by others, and revise their readiness or willingness to speak. A simulation framework that represents only observable turn exchange would therefore miss much of the process through which discussion evolves. \modelname{} addresses this limitation by making the latent reasoning process underlying public expression analytically visible. At each interval, agents update their internal states based on the shared dialogue history and their own evolving evaluations, while public speech remains temporally constrained and globally coherent. This design treats speech as the observable outcome of continuous internal updating, allowing opinion dynamics and communication strategies to emerge from evolving evaluation rather than from mechanical turn succession alone.

Recent work has begun to clarify how generative AI may be used to support democratic deliberation. One line of research treats AI as an intervention within human discussion. For example, \citet{argyleLeveragingAIDemocratic2023} show that real-time AI-generated rephrasing suggestions can improve online political conversations by helping participants express disagreement in ways that convey listening, validation, and respect. Relatedly, \citet{tesslerAICanHelp2024} develop the Habermas Machine, an AI mediator that synthesizes individual opinions and critiques into group statements designed to capture common ground. Their findings suggest that AI-mediated deliberation can produce statements preferred to those written by human mediators, reduce group division, and incorporate minority critiques into revised statements. Together, these studies demonstrate that AI can function as deliberative infrastructure: it can help human participants communicate more constructively, feel better understood, and identify points of shared agreement.

A second line of work moves from AI-assisted deliberation to AI-generated or AI-guided deliberative simulation. \citet{rountreeCaseUsingGenerative2026} make a compelling case for using generative AI to run deliberation simulations that complement, rather than replace, human judgment. They frame such simulations as ``deliberation-making'' tools rather than decision-making shortcuts, with potential applications in facilitator training, time-sensitive policy consultation, classroom deliberation, and theory development. \citet{ashkinazePluralsSystemGuiding2025} offer a more concrete multi-agent system in this direction. Their Plurals system organizes persona-based LLM agents into customizable deliberative structures, with moderators synthesizing the resulting communication. This work demonstrates the value of moving beyond a single ``view from nowhere'' and instead using simulated social ensembles to represent diverse perspectives under different rules of interaction.

Despite these advances, existing approaches tend to emphasize either the improvement of human deliberation or the generation of deliberative outputs. AI-assisted systems focus on helping human participants phrase messages, synthesize opinions, or find common ground. Simulation-oriented systems often generate transcript-like exchanges among personas or produce summaries, proposals, rankings, and tradeoffs. These contributions are important, but they leave a core communication problem undertheorized: how participation emerges over the course of an unfolding public discussion. In many deliberative settings, especially town hall meetings, the central process is not simply what participants say once selected to speak. Participants also listen, interpret prior remarks, revise their beliefs, reassess the relevance of the issue, monitor the interactional climate, and decide whether the moment gives them a reason to enter the discussion. Much of this process occurs while participants remain silent.

Our study addresses this gap by developing a simulation framework that models deliberation as an unfolding communication process. Rather than treating agents only as sources of final opinions or generated turns, our design gives agents interval-level internal-state updating. After each public remark, agents who do not speak still update their internal states, including their understanding of the issue, perceived disagreement, concern, relevance, and readiness to speak. Speaking behavior then emerges from these changing internal states rather than from a simple turn-generation procedure. This design is especially suited to town-hall settings, where silence, listening, speaking readiness, and turn entry are central features of public participation.

In this sense, our study contributes to the emerging literature on AI and deliberation in two ways. Conceptually, it shifts attention from AI-assisted deliberation and deliberative output generation to the simulation of deliberative communication. Methodologically, it proposes a mechanism-sensitive architecture for representing latent processes that are difficult to observe directly in human deliberation. By modeling how agents listen, update, remain silent, and eventually decide whether to speak, the framework allows LLM-based simulation to be used not only to produce plausible deliberative talk, but also to examine the communicative mechanisms through which public discussion unfolds.

\section{Additional Illustrations of Baseline Frameworks}
\label{app:baseline_figures}

This section provides schematic illustrations of the two baseline interaction paradigms discussed in Section~\ref{sec:baseline_frameworks}. 

\begin{figure*}[ht]

\begin{center}
\centerline{\includegraphics[width=0.65\textwidth]{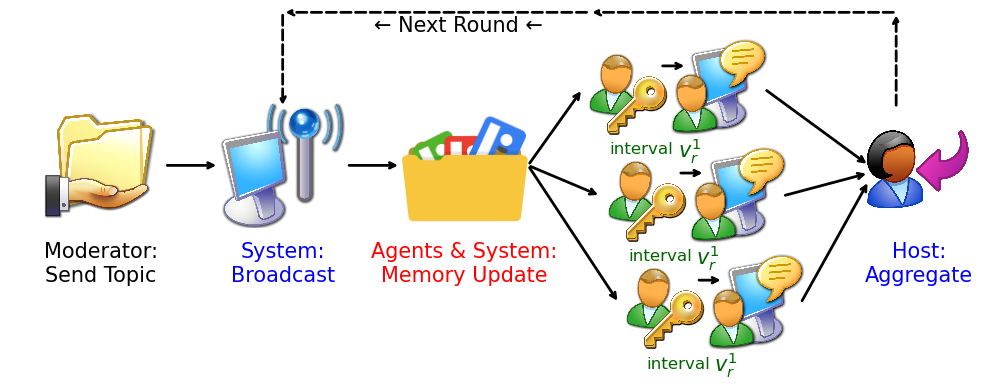}}

\caption{Framework of \texttt{Hierarchical} Multi-agent System.}
\label{HierarchicalSimulator}
\end{center}

\end{figure*}

\begin{figure*}[ht]

\begin{center}
\centerline{\includegraphics[width=0.95\textwidth]{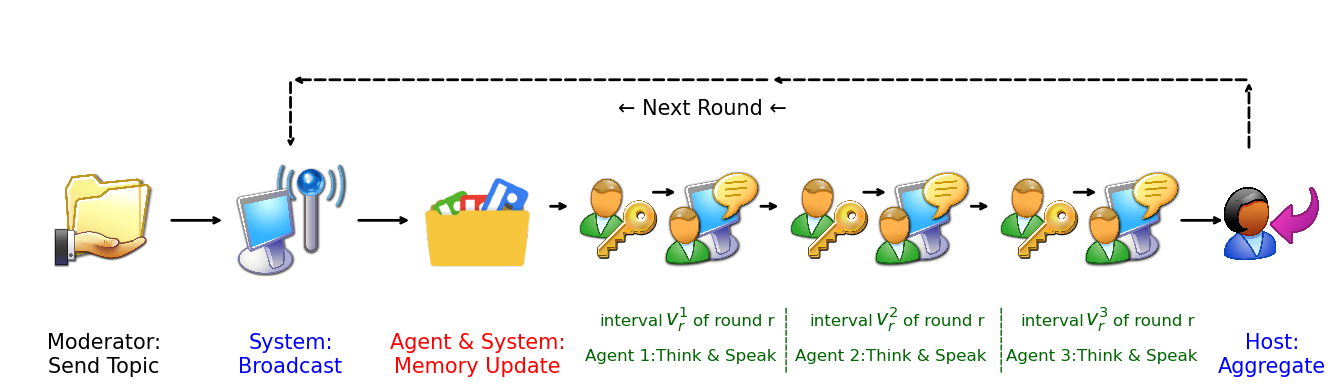}}

\caption{Framework of \texttt{Sequential} Multi-agent System.}
\label{SequentialSimulator}
\end{center}
\end{figure*}

Figure~\ref{HierarchicalSimulator} illustrates the hierarchical aggregation setting, where agents generate outputs in parallel based on previous rounds and the host aggregates their utterances. Figure~\ref{SequentialSimulator} illustrates the sequential turn-taking setting, where agents are called one by one according to a predefined order, and each agent observes the prior public utterances before producing its own response.

\section{Detailed Analysis and Results}
\label{app:result}

\subsection{Construction and Interpretation of Internal-State Indices}

The main analysis uses two composite indices to summarize agents' interval-level internal evaluations. The first index, dissonance-related appraisal, represents the cognitive-dissonance mechanism motivating the town hall simulation. It combines three structured indicators generated by the agents: perceived inconsistency, dissonance tension, and motivation to reduce dissonance. The second index, perceived silence pressure, represents the spiral-of-silence and self-censorship mechanisms. It combines an unfavorable perceived opinion climate indicator and perceived isolation risk.

The unfavorable perceived opinion climate variable was constructed by reverse-coding and rescaling the original perceived opinion climate score so that larger values represented a more unfavorable climate for the agent's own position. We then averaged the corresponding indicators to obtain interval-level composite scores. The composite construction was empirically supported by high internal coherence, as reported in the main text.

These indices should be interpreted as structured process traces generated by the simulation, not as validated psychometric measures of human psychological states. Their purpose is to make agents' latent evaluations analyzable during the unfolding discussion. This design allows the analysis to move beyond transcript-only simulation outputs and examine whether theory-relevant appraisals vary systematically over time and across experimental conditions.

\subsection{Mixed-Effects Models for Internal-State Dynamics}

To analyze the temporal dynamics of dissonance and silence pressure, we estimated two linear mixed-effects models. Each model used one internal-state index as the outcome and included centered interval, persona ecology, turn-allocation rule, Force Speak setting, and memory mode as fixed effects. To test whether temporal trajectories differed by simulation design, we also included interactions between centered interval and each design factor. Random intercepts were included for both simulation run and agent, accounting for repeated observations within runs and within agents.

For agent $i$ in run $r$ at interval $t$, the general model was:

\[
\begin{aligned}
Y_{irt} =\;& \beta_0 
+ \beta_1 \text{Interval}_{t} 
+ \beta_2 \text{PE}_{r}
+ \beta_3 \text{TT}_{r} \\
&+ \beta_4 \text{FS}_{r} 
+ \beta_5 \text{MM}_{r} \\
&+ \beta_6(\text{Interval}_{t} \times \text{PE}_{r}) 
+ \beta_7(\text{Interval}_{t} \times \text{TT}_{r}) \\
&+ \beta_8(\text{Interval}_{t} \times \text{FS}_{r})
+ \beta_9(\text{Interval}_{t} \times \text{MM}_{r}) \\
&+ u_r + v_i + \epsilon_{irt}.
\end{aligned}
\]

Here, $Y_{irt}$ denotes either the dissonance index or the silence-pressure index, $u_r$ is a run-level random intercept, $v_i$ is an agent-level random intercept, and $\epsilon_{irt}$ is the residual. This specification separates average condition differences from condition-specific changes over the course of the discussion.

The main text reports the central coefficients for both internal-state models. Additional inspection of the models shows that memory mode did not produce significant main effects on the average dissonance level, whereas summary memory increased the average level of silence pressure compared with the \texttt{latest} memory mode. The interval-by-Force Speak interaction was not significant in either model, suggesting that Force Speak affected average levels more than temporal slopes.

\subsection{Interpretation of Internal-State Dynamics}

The internal-state analysis addresses both RQ1 and RQ3. For RQ1, the results show that \modelname{} produces interval-level traces that can be transformed into coherent, interpretable, and statistically analyzable process indicators. These traces reveal aspects of the simulation that would not be available from transcript-only models, because agents who do not speak still produce internal evaluations.

For RQ3, the results show that internal states are shaped by simulation design. Persona ecology, turn-allocation rule, Force Speak settings, and memory mechanisms affect either average levels, temporal trajectories, or both. The most stable pattern concerns turn allocation: the willing mode increased both dissonance and silence pressure and also amplified their growth over time. This means that self-selection does more than change who speaks; it also changes the internal experience of the discussion among agents.

\subsection{Logistic Mixed-Effects Model for Willingness to Speak}

After establishing that internal states were coherent and dynamic, we examined whether they predicted speaking intention. This analysis directly addresses RQ2, which asks whether public expression emerges from internal evaluation rather than from fixed turn order or immediate reactive response alone.

The outcome was whether an agent wanted to speak at a given interval. The logistic mixed-effects model included the dissonance index, the silence-pressure index, centered interval, persona ecology, turn-allocation rule, Force Speak setting, and memory mode. To test whether internal-state effects differed between turn-allocation structures, we included interactions between turn-allocation rule and each internal-state index. Random intercepts were included for simulation run and agent. Because the internal-state indices were originally bounded between 0 and 1, they were rescaled so that each coefficient corresponded to a .10-unit increase in the original scale.

The main text reports the central internal-state effects, turn-allocation interactions, and major design-factor effects. Force Speak and memory mode did not show significant direct effects after internal-state indices and design factors were included. Overall, this model supports the claim that speaking intention is not merely an artifact of the simulation procedure. Agents' willingness to speak was organized around their evolving internal evaluations while listening to the discussion.

\subsection{Modeling the Transition from Intention to Public Expression}

The final analysis examined whether agents who wanted to speak were actually allowed to speak. The analysis was restricted to agent-interval observations in which the agent had formed a speaking intention. The outcome was whether that intention became the publicly selected utterance in the corresponding interval. This analysis therefore focuses on the second stage of expression: the movement from internal speaking intention to realized public speech.

The planned mixed-effects model produced a singular fit, so we used a binomial logistic regression as a descriptive follow-up. The predictors were the dissonance index, silence-pressure index, centered interval, persona ecology, turn-allocation rule, Force Speak setting, and memory mode. The two internal-state indices were again rescaled so that each coefficient represented a .10-unit increase in the original index. Interactions between turn-allocation rule and each internal-state index were also included.

The main text reports the central results from this second-stage model. Force Speak and memory mode were not significantly associated with being allowed to speak. Together, the willingness and public-expression models support a two-stage interpretation. In the first stage, internal evaluations shape whether agents develop a desire or intention to speak. In the second stage, once speaking intention exists, whether that intention becomes public expression is governed primarily by turn-allocation rules. This distinction is central to the framework because it separates latent communicative motivation from observable speech.

\end{document}